\documentclass{ecai} 

\usepackage{latexsym}
\usepackage{amssymb}
\usepackage{amsmath}
\usepackage{amsthm}
\usepackage{booktabs}
\usepackage{enumitem}
\usepackage{graphicx}
\usepackage{color}
\usepackage{algorithmicx}
\usepackage{algpseudocode}
\usepackage[linesnumbered,ruled,vlined]{algorithm2e}

\newcommand{\BibTeX}{B\kern-.05em{\sc i\kern-.025em b}\kern-.08em\TeX}

\begin{document}


\begin{frontmatter}


\paperid{1797} 


\title{FGML-DG: Feynman-Inspired Cognitive Science Paradigm for Cross-Domain Medical Image Segmentation}


\author[A]{\fnms{Yucheng}~\snm{Song}}
\author[A]{\fnms{Chenxi}~\snm{Li}}
\author[A]{\fnms{Haokang}~\snm{Ding}}
\author[B]{\fnms{Zhining}~\snm{Liao}} 
\author[A]{\fnms{Zhifang}~\snm{Liao}\thanks{Corresponding Author. Email: zfliao@csu.edu.cn}} 

\address[A]{School of Computer Science and Engineering, Central South University, Changsha, China}
\address[B]{School of Health \& Wellbeing, University of Glasgow, UK}


\begin{abstract}
In medical image segmentation across multiple modalities (e.g., MRI, CT, etc.) and heterogeneous data sources (e.g., different hospitals and devices), Domain Generalization (DG) remains a critical challenge in AI-driven healthcare. This challenge primarily arises from domain shifts, imaging variations, and patient diversity, which often lead to degraded model performance in unseen domains. To address these limitations, we identify key issues in existing methods, including insufficient simplification of complex style features, inadequate reuse of domain knowledge, and a lack of feedback-driven optimization. To tackle these problems, inspired by Feynman’s learning techniques in educational psychology, this paper introduces a cognitive science-inspired meta-learning paradigm for medical image domain generalization segmentation. We propose, for the first time, a cognitive-inspired Feynman-Guided Meta-Learning framework for medical image domain generalization segmentation (FGML-DG), which mimics human cognitive learning processes to enhance model learning and knowledge transfer. Specifically, we first leverage the ‘concept understanding’ principle from Feynman’s learning method to simplify complex features across domains into style information statistics, achieving precise style feature alignment. Second, we design a meta-style memory and recall method (MetaStyle) to emulate the human memory system’s utilization of past knowledge. Finally, we incorporate a Feedback-Driven Re-Training strategy (FDRT), which mimics Feynman’s emphasis on targeted relearning, enabling the model to dynamically adjust learning focus based on prediction errors. Experimental results demonstrate that our method outperforms other existing domain generalization approaches on two challenging medical image domain generalization tasks.
\end{abstract}

\end{frontmatter}


\section{Introduction}

Generalized segmentation of medical images across multiple modalities (such as MRI, CT, and others) and heterogeneous data sources (such as different hospitals and devices) remains a critical challenge in artificial intelligence (AI)-driven healthcare \cite{yoon2024domain,zhou2022device,zhou2024rethinking,zheng2024advst}. Variations in imaging principles, gray-scale distributions, and texture features, combined with inter-patient variability and disease diversity, lead to significant domain shift, which hinders the ability of segmentation models to extract universal patterns \cite{yi2024hallucinated}. Traditional methods often fail to adapt to unseen domains, thereby limiting their clinical applicability \cite{wang2025partial}. Drawing inspiration from human cognitive mechanisms, this study integrates AI with educational psychology by incorporating the Feynman learning technique into meta-learning approaches, thereby constructing a domain generalization segmentation framework for medical imaging. This interdisciplinary perspective aims to enhance domain generalization capabilities, enabling the model to emulate human learning and knowledge transfer in complex and dynamic medical imaging scenarios.

\begin{figure}[ht]
  \centering
  \includegraphics[width=0.9\linewidth]{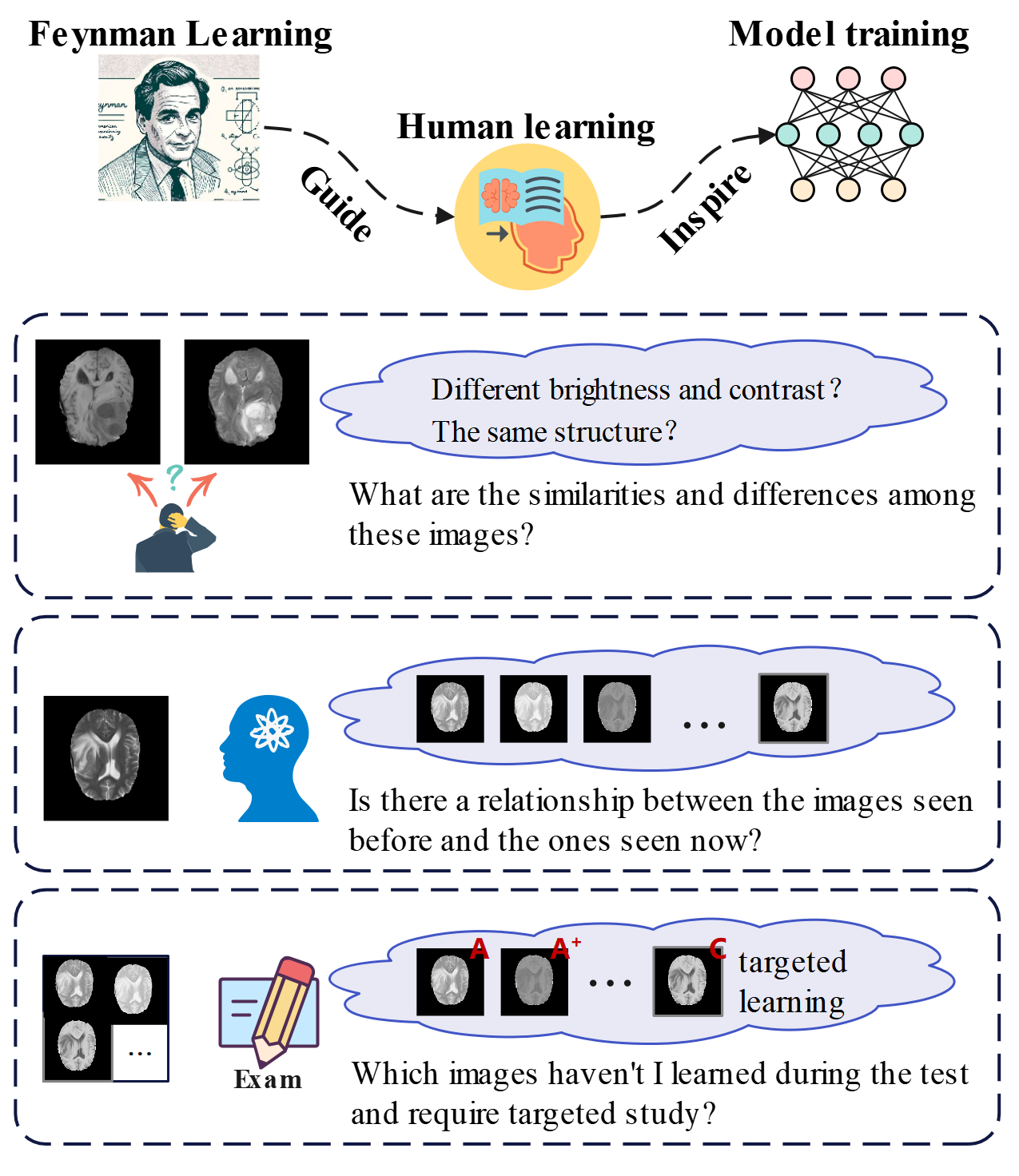}
  \caption{Illustration of the motivation for FGML-DG, inspired by human cognitive mechanisms through the Feynman learning technique, including: (1) understanding and simplifying style concepts in medical images, (2) reusing and memorizing existing style knowledge, and (3) conducting targeted feedback learning on errors during the learning process}
  \label{fig1}
\end{figure}

Existing meta-learning methods have made certain advances in domain generalization (DG) for medical imaging, primarily by employing a “learn to learn” strategy to enhance the model’s adaptability to cross-domain data \cite{khandelwal2020domain,qin2023bi,shu2021open,zhong2022meta}. For example, some methods capture domain-invariant features by designing shared feature extraction networks, or leverage knowledge from gradient information to support learning in new tasks \cite{singh2021metamed,liu2021semi,sicilia2021multi,wang2024domain}. However, these methods still face significant challenges when handling heterogeneity across modalities (such as MRI and CT) and institutional data sources. Inspired by the Feynman learning technique, we re-examine these issues from three unique perspectives, as shown in Figure \ref{fig1}: first, from the perspective of understanding complex concepts in the Feynman learning technique, it is necessary to deeply understand and effectively simplify complex stylistic features (such as imaging principles, gray-scale distributions, and texture features), thereby improving the model’s cross-domain knowledge transfer efficacy; second, by analogy to the note-taking practice in the Feynman learning technique that reinforces memory, existing methods overlook the contribution of known domain knowledge in simulating features of unseen domains, thereby limiting the model’s knowledge reuse capabilities; finally, based on the idea in the Feynman learning technique of targeted learning through feedback, existing methods lack an effective feedback mechanism, preventing dynamic optimization of learning strategies based on model performance. The existence of these problems severely constrains the generalization performance of meta-learning models in complex medical imaging data.

To address the aforementioned issues, we propose a cognitively inspired Feynman-Guided Meta-Learning framework for domain generalization in medical image segmentation (FGML-DG). First, we design a meta-style knowledge alignment method that, inspired by the ‘concept understanding’ aspect of the Feynman learning technique, simplifies complex features from cross-domain data into style information statistics (such as mean and variance), and achieves precise alignment of style features using contrastive learning and consistency learning. Second, we design a meta-style memory and review method (MetaStyle) to store and replay domain-invariant style statistics, simulating unseen domains to enable proactive knowledge reuse. Finally, we introduce a Feedback-Driven Re-Training strategy (FDRT) that emulates the targeted re-learning emphasized in the Feynman learning technique, allowing the model to dynamically adjust its learning focus based on prediction errors. These cognitively inspired designs provide novel approaches to overcoming the limitations of existing methods, enabling the model to perform exceptionally in cross-domain generalization tasks by mimicking human-like learning processes. We summarize our main contributions as follows:
\begin{itemize}
    \item We are the first to integrate the Feynman learning technique into meta-learning for medical imaging, proposing a cognitively inspired meta-learning framework that significantly enhances the model’s segmentation capabilities in cross-domain generalization.
    \item We propose a systematic framework that addresses the challenges of knowledge alignment, retention, and optimization. Specifically, we design a meta-style knowledge alignment method to achieve precise feature alignment across domains; a meta-style memory and review method (MetaStyle) to enhance the simulation of unseen domain features and improve knowledge reuse capabilities; and a FDRT strategy to dynamically optimize the model’s performance.
    \item We conduct experiments on multiple medical image segmentation datasets. The results demonstrate that FGML-DG achieved excellent generalization performance, advancing AI towards cognitively inspired adaptability.
\end{itemize}
\section{Relate Work}
\subsection{Meta-Learning for DG in Medical Imaging}
Domain generalization (DG) in medical image analysis is a long-standing challenge, particularly in applications involving cross-modalities (such as MRI, CT, PET) and cross-institutional data. Meta-learning has made certain advances in enhancing generalization capabilities in the medical imaging domain, primarily through a “learn to learn” strategy to improve the model’s adaptability to unknown domains \cite{khandelwal2020domain}. For example, Liu et al. \cite{liu2020shape} utilized virtual meta-training and meta-testing to explicitly simulate domain shifts. Li et al. \cite{li2022domain} proposed a meta-learning-based task augmentation method that simulates domain drift to improve the model’s generalization in new domains. Alternatively, gradient-based meta-learning methods simulate domain drift during optimization and enhance feature semantic structure stability through regularization techniques to improve generalization \cite{dou2019domain}. Despite the progress of meta-learning in domain generalization for medical imaging, existing methods still face significant challenges. First, these methods often rely on explicit simulation of domain shifts or task augmentation, but they struggle to capture domain-specific style information (such as imaging principles, gray-scale distributions, and texture features) when dealing with the complex heterogeneity of cross-modalities (e.g., MRI, CT) and cross-institutional data, resulting in limited knowledge transfer effectiveness. Second, existing methods are insufficient in dynamically adapting to unseen domain features and optimizing model feedback mechanisms, which limits the model’s generalization ability in complex medical imaging data. Therefore, how to introduce new cognitive mechanisms to improve the performance of meta-learning in domain generalization for medical imaging remains a problem worthy of in-depth exploration.

\subsection{Cognitively Inspired AI Learning Paradigm}
Cognitive science, as a discipline that studies human learning and information processing mechanisms, provides important theoretical guidance for the optimization of artificial intelligence models \cite{agrawal2023advancing}. In recent years, researchers have attempted to draw from cognitive science perspectives, leveraging the characteristics of human learning processes to optimize the training methods of deep learning models, thereby enhancing model performance \cite{jiao2024brain,stettler2018using,lin2021training}. For instance, inspired by theories of memory systems in cognitive psychology, some studies have proposed neural network architectures that integrate short-term and long-term memory, to simulate human mechanisms for storing and retrieving knowledge during learning \cite{xiang2020cslm,nachstedt2015towards}. Zhao et al. \cite{zhao2017deep} introduced a deep active learning method based on human cognitive processes, which improves the training efficiency of deep learning models by progressively acquiring additional knowledge. Veksler et al. \cite{veksler2022symbolic} drew upon psychological theories to propose a Symbolic Deep-learning Network (SDN), addressing issues in Deep Neural Networks (DNNs) related to learning efficiency, catastrophic forgetting, and interpretability. Thus, AI learning paradigms inspired by cognitive science offer new insights for various domains of deep learning. However, no existing research has combined cognitive science principles, particularly the Feynman learning technique, with meta-learning frameworks to optimize domain generalization capabilities in medical image analysis. Therefore, we propose a Feynman-Guided Meta-Learning framework (FGML-DG) to simulate human learning processes, thereby enhancing the cross-domain adaptability of medical image models.


\section{Methods}
\begin{figure*}[ht]
  \centering
  \includegraphics[width=0.8\linewidth]{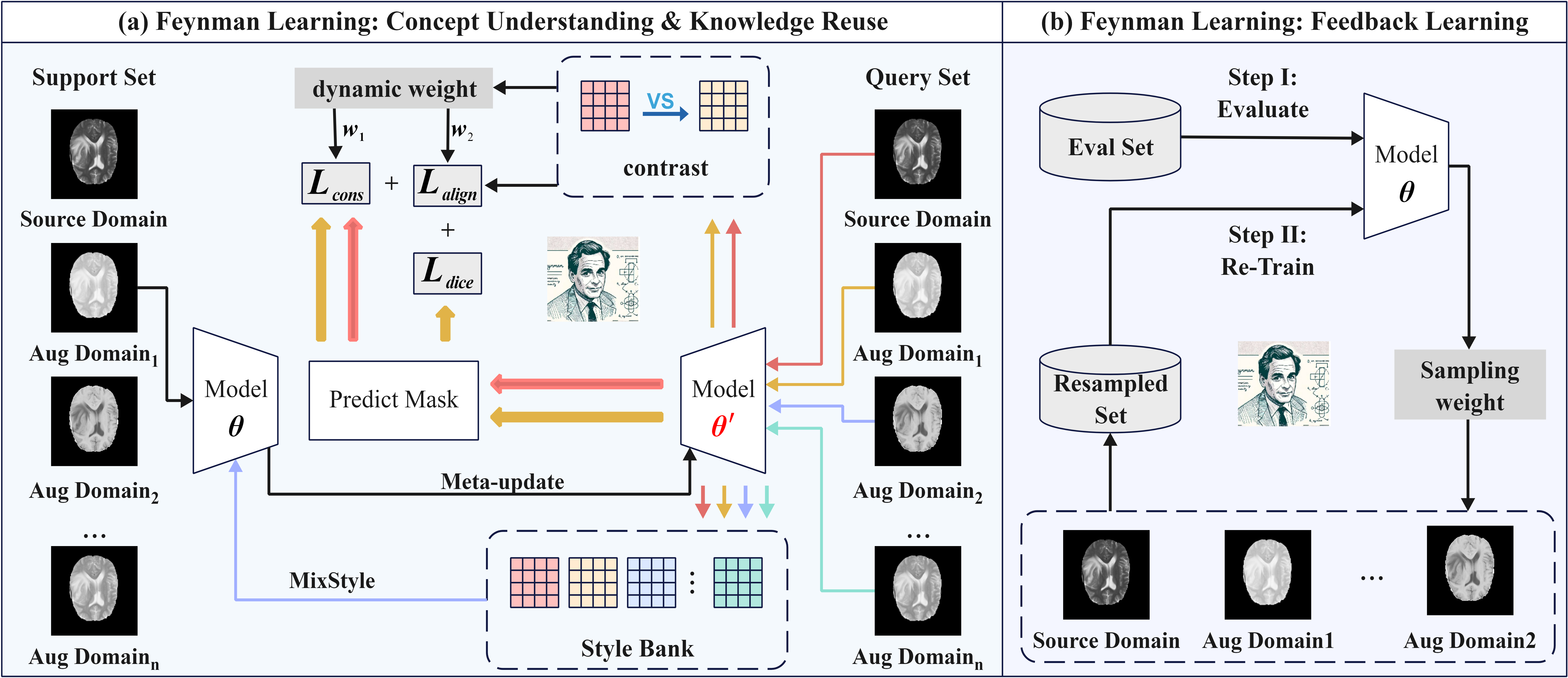}
  \caption{Overview of the FGML-DG framework. (a) In the meta-learning stage, we employ meta-style knowledge alignment methods and meta-style memory and review methods, inspired by the Feynman learning technique, to train the model. (b) Drawing from the feedback-targeted learning aspect of Feynman learning, we designed a feedback-driven retraining strategy that allows the model to dynamically adjust its learning focus based on prediction errors.}
  \label{fig2}
\end{figure*}

In this section, we first present a description of the problem. Subsequently, we provide a detailed explanation of our method. Figure \ref{fig2} offers an overview of the method framework.

\subsection{Preliminaries}
\begin{figure}[ht]
  \centering
  \includegraphics[width=0.8\linewidth]{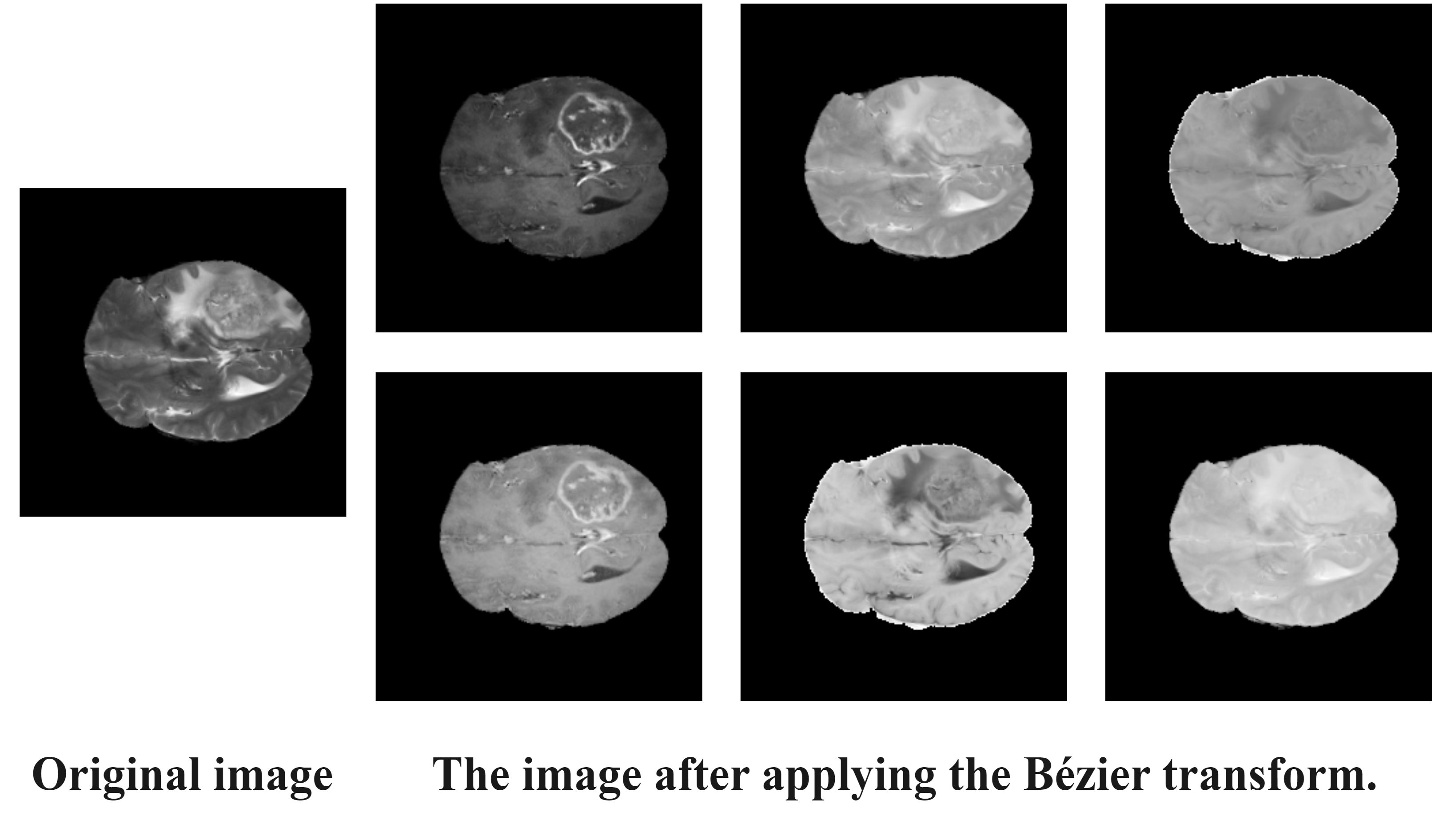}
  \caption{By employing Bezier transformations, style divergence can be facilitated from the source domain to different style domains.}
  \label{fig}
\end{figure}

In the medical image domain generalization segmentation task, we assume the existence of a source domain \(D^S\) and a target domain \(D^T\), where \(D^S\) can be defined as \(\{x_i, y_i\}_{i=1}^N\), with \(x_i\) representing the i-th sample in the source domain, \(y_i\) representing the corresponding segmentation mask for that sample, and \(N\) being the total number of samples. We define a model parameterized by \(\Theta\) to address the specified task. The goal of DG is to train \(\Theta\) on \(D^S\) such that it can generalize to \(D^T\). To achieve this, we first employ a simple and effective data augmentation strategy \cite{zhou2022generalizable} based on nonlinear transformation \(B(\cdot)\), which generates an augmented domain \(B(D^S) = \{D^0, \dots, D^A\}\) to simulate the potential target domain distribution. This is because the flexibility of \(B(\cdot)\) allows for efficient exploration of the continuous style space, thereby simulating real train-test domain shifts, enabling the training of a model over many iterations to achieve good generalization on \(D^T\). We present examples of images generated using \(B(\cdot)\), as illustrated in Figure \ref{fig}.

\subsection{Meta-Style Knowledge Alignment Method} 
Recent style transfer studies have shown that style information from different domains is preserved in the lower layers of CNNs through instance-level feature statistics \cite{huang2017arbitrary,zhou2024mixstyle,chen2022maxstyle,liu2024stydesty}. Therefore, we first draw from the conceptual knowledge understanding perspective of Feynman learning law and propose a knowledge alignment method based on style concepts, aimed at reducing style differences between different domains. Specifically, for the \(a\)-th augmented domain, we align the statistical characteristics of \(D^S\) and \(D^a\) through features in the intermediate layers of the network. Assuming that input data \(x\) is processed by the model's convolutional layer to extract shallow features \(F = f(x) \in \mathbb{R}^{B \times C \times H \times W}\), the style information of images within the domain is described using the mean \(\mu\) and standard deviation \(\sigma\):

\begin{equation}
    \mu = \frac{1}{H \cdot W} \sum_{h=1}^{H} \sum_{w=1}^{W} F_{c, h, w}
    \label{eq1}
\end{equation}

\begin{equation}
    \sigma = \sqrt{\frac{1}{H \cdot W} \sum_{h=1}^{H} \sum_{w=1}^{W} (F_{c, h, w} - \mu)^2}
    \label{eq2}
\end{equation}

where \(B\) is the batch size, \(C\) is the number of channels in the feature map, and \(H, W\) are the spatial dimensions of the feature map. Based on this, we design a style alignment loss based on contrastive learning, with the goal of transforming the shallow features of the source domain to the style distribution of the augmented domain, enabling the model to adapt to changes in distribution across different modalities.

\begin{equation}
\begin{aligned}
L_{align} &= \frac{1}{n} \sum_{i = 1}^{n} \Bigg[ l_i \times D(x_i, x_i^a)^2 \\
& \quad + (1 - l_i) \times (\max(0, m - D(x_i, x_i^a)))^2 \Bigg]
\end{aligned}
\label{eqalign}
\end{equation}

Where, \(x_i\) and \(x_i^a\) are the normalized feature vectors from the source domain and the \(a\)-th augmented domain, respectively; \(l_i\) denotes class similarity, with \(l_i \in \{0, 1\}\); \(D(\cdot)\) represents the Euclidean distance; and \(m\) is the margin parameter.

Additionally, to enforce consistency between the predictive outputs of the source domain and the augmented domain, thereby reducing the impact of domain shift on model performance, we introduce a consistency loss function.

\begin{equation}
    L_{cons} = \frac{1}{n} \sum_{i=1}^{n} \left\| p(x_i) - p(x_i^a) \right\|_2^2
    \label{cons}
\end{equation}

Furthermore, to avoid over-alignment in similar domains and insufficient alignment in dissimilar domains, we incorporate an inter-domain offset into the style alignment loss and design a dynamic adjustment mechanism to determine the loss weights. This enhances the model's adaptability to different domains and uses logarithmic compression to increase sensitivity to inter-domain style differences.

\begin{equation}
\begin{aligned}
\Delta_{style} &= \ln \left( 1 + \left( \frac{1}{N} \sum_{i=1}^{N} |\mu_i - \mu_i^a| \right) \times s \right) \\
& \quad + \ln \left( 1 + \left( \frac{1}{N} \sum_{i=1}^{N} |\sigma_i - \sigma_i^a| \right) \times s \right)
\end{aligned}
\end{equation}

\begin{equation}
    w = 1 - \exp(-\Delta_{style})
\end{equation}

where \(s\) is a sensitivity hyperparameter, and the dynamic weight \(w\) is designed to range within \([0, 1)\), reflecting the magnitude of style differences and used to adjust the intensity of the training process.

\begin{algorithm}
    \caption{MetaStyle}
    \label{alg1}
    \KwIn{Domain list $T$, Dataset $[D_0, \ldots, D_T]$, Style Bank list $P$, Hyperparameter $\gamma$ and $\beta$, Model parameters $\theta$}
    \For{$t = 0, \ldots, T$}{
        \If{$t > 0$}{
            $S_{old} \leftarrow$ load\_style\_bank($P[t - 1]$)
        }
        \For {Sample data $x_{spt}$ from $D[t]$}{
        $\mu_x, \sigma_x \leftarrow$ mean($x_{spt}$), std($x_{spt}$) \\
        \If{$t = 0$}{
            save\_style\_bank($S, P[t]$)
        }
        $\mu_{mix}, \sigma_{mix} \leftarrow S_{old}$ \algorithmiccomment{Review historical style}\\
        \textbf{Meta-Train}: \text{Grads} $\nabla_{\theta} = F'_{\theta}(x_{spt}; \mu_{mix}, \sigma_{mix}; \theta)$\\
        Updated parameters $\theta' = \theta - \gamma\nabla_{\theta}$\\
        }
        \For {sample data $x_{qry}$ from $D[t + 1]$} {
        $\mu_x, \sigma_x \leftarrow$ mean($x_{spt}$), std($x_{spt}$) \\
        \If{$t \neq 0$}{
            save\_style\_bank($S, P[t]$)
        }
        \textbf{Meta-Test}: \text{Grads} $\nabla_{\theta'} = F'_{\theta'}(x_{qry}; \theta')$\\
        Updated parameters $\theta' = \theta' - \beta\nabla_{\theta'}$
        }
    }
        \textbf{End}
\end{algorithm}

\subsection{Meta-Style Memory and Review Method}
In the FGML-DG framework, we design the MetaStyle memory and recall method (MetaStyle) to mimic the human memory system's utilization of past knowledge. This method leverages the meta-learning paradigm, treating the current domain data as the support set and the next domain data as the query set, and dynamically stores and recalls instance-level feature statistics through a style bank module. Specifically, during the meta-training phase, we load the style statistics from the previous domain as prior knowledge and mix them with shallow network features; during the meta-testing phase, we save the shallow network statistics of the current task for future recall. The storage method allows for the dynamic accumulation of statistics from multiple instances and, during recall, through weighted averaging mixing:

\begin{equation}
    \mu_{mix} = \alpha\mu_{current} + (1 - \alpha)\mu_{old}
    \label{eq7}
\end{equation}

\begin{equation}
    \quad \sigma_{mix} = \alpha\sigma_{current} + (1 - \alpha)\sigma_{old}
    \label{eq8}
\end{equation}

where \(\alpha\) is the mixing coefficient, and \(\mu\) and \(\sigma\) represent the mean and standard deviation, respectively. Then, in the corresponding network layer, the mixed mean and standard deviation are used to normalize the current feature map, thereby achieving the recall of style knowledge, with the specific formula being:

\begin{equation}
    x_{output} = \left(\frac{x - \mu_x}{\sigma_x + \epsilon}\right) \times \sigma_{mix} + \mu_{mix}
    \label{eq9}
\end{equation}

In Algorithm \ref{alg1}, we provide a detailed description of the specific process for MetaStyle.

\subsection{Feedback-Driven Re-Training Strategy}

In the Feynman learning technique, output serves as the critical link for assessing whether learning has been successful, while feedback provides an evaluation of the output's effectiveness. Within the FGML-DG framework, the FDRT emphasizes identifying weaknesses through output feedback and implementing targeted learning to address them. As illustrated in Algorithm \ref{alg2}, this strategy is designed to tackle the problem of insufficient adaptation to unseen domains in medical image domain generalization tasks. Specifically, we leverage prediction feedback from the validation set to quantify the performance disparities of the model across different domains. Subsequently, we devise a weighted sampling method to select data instances where the model's predictive performance is suboptimal, akin to knowledge points that a learner has not yet mastered.

\begin{equation}
    Domain_i = 1 - \exp(\lg(m_i))
\end{equation}

Where, \( m_i \) represents the validation result on the i-th domain, and \( Domain_i \) denotes the sampling proportion during FDRT retraining. By iteratively repeating this process, the model emulates a learner employing the Feynman learning technique, progressively deepening its understanding of knowledge and thereby enhancing its generalization capability on data from diverse domains.

\begin{algorithm}
    \caption{Feedback-Driven Re-Training Strategy.}
    \label{alg2}
    \KwIn{Domain list $T$, Dateset $[D_0, \ldots, D_T]$, Validation set $[V_0, \ldots, V_T]$, Hyperparameters $\eta$, Model parameters $\theta$}
    \While{Not Converged}{
        Meta-Learning([$D_0, \ldots, D_T$])\\
        $M = \{m_0, \ldots, m_T\} \leftarrow$ Eval([$V_0, \ldots, V_T$]; $\theta$)\\
        $G \leftarrow$ calculate\_Gap($M$)\\
        \For{$t = 0, \ldots, T$}{
            \For {Sample data $x_{retrain}$ from $D[t] * G[t]$}{
            \textbf{FDRT-Train}: Grads $\nabla_{\theta} = F'_{\theta}(x_{retrain}; \theta)$\\
            Updated parameters $\theta = \theta - \eta\nabla_{\theta}$
            }
            }
        }
    \textbf{End}
\end{algorithm}

\subsection{Loss Function}
In the FGML-DG framework, the loss function integrates style alignment loss, consistency loss, and the main task segmentation loss to enhance cross-domain adaptation and generalization capabilities. Drawing inspiration from the Feynman learning technique, our loss function not only focuses on the performance of the segmentation task but also incorporates auxiliary losses (style and consistency) to simulate the feedback and knowledge alignment processes inherent in human learning. The auxiliary loss is defined as:

\begin{equation}
    L_{aux} = (1 - w)L_{cons} + wL_{style}
\end{equation}

For the segmentation task, we utilize Dice loss, and the total loss function is given by:

\begin{equation}
    L_{total} = \lambda L_{aux} + (1 - \lambda)L_{dice}
\end{equation}

\section{Experiments}
\subsection{Datasets}
We introduce two datasets (namely the BraTS dataset \cite{menze2014multimodal} and the Abdominal Multi-Organ dataset \cite{kavur2021chaos,landman2015miccai}) for evaluation. The Cross-Modality Brain Tumor Segmentation Challenge 2018 dataset (BraTS) includes 210 cases of high-grade gliomas and 75 cases of low-grade gliomas, with each case comprising MRI images from four modalities: T2, Flair, T1, and T1CE. For fair comparison, we follow the settings of Zhou et al. \cite{zhou2022generalizable}, using T1CE and T2 as the source domain for experiments and validating the results on the other three domains.

The Abdominal Multi-Organ dataset encompasses the CHAOS \cite{kavur2021chaos} and SABSCT \cite{landman2015miccai} datasets. CHAOS is an abdominal MRI dataset that provides annotations for multiple abdominal organs. SABSCT contains abdominal imaging data extracted from CT scans, with precise annotations for several major organs. On this dataset, we separately use MRI and CT as the source domains for experiments. Additionally, the training data from the source domain is split into training and validation sets in a 7:3 ratio.
\subsection{Implement detail}

We adopted U-Net \cite{ronneberger2015u} as our meta-learning segmentation backbone network. All experiments were conducted in an environment using Python 3.9 and PyTorch 2.4 on an Ubuntu 22.04 system, with four NVIDIA RTX 4090 GPUs. For model training, we set the number of epochs to 200, with a batch size of 8, and employed the Adam optimizer. In the meta-learning process, the learning rate for meta-training was set to 0.01, and for meta-testing to 0.005. For the feedback-driven retraining strategy, the initial learning rate was 0.01, with the number of training epochs set to 100. Additionally, we implemented a learning rate decay strategy, reducing the learning rate by a factor of 0.5 every 20 epochs. All results are reported as the average over five different random seeds to ensure the stability of our method.

We utilized two commonly used evaluation metrics: the Dice coefficient (Dice) and the Hausdorff distance (HD). The Dice coefficient measures the overlap between the prediction and the ground truth, with higher values indicating better segmentation performance. The Hausdorff distance is defined between two sets in a metric space, and lower values indicate better performance.

\subsection{Comparison Experiment}
We evaluated the performance of our proposed FGML-DG framework on two challenging medical image segmentation datasets and compared it with other state-of-the-art domain generalization methods. Specifically, we compared our method with baseline approaches, including Fed-DG \cite{liu2021feddg}, MixStyle \cite{zhou2021domain}, CSDG \cite{ouyang2022causality}, SADN \cite{zhou2022generalizable}, EGSDG \cite{gutrain}, Cutout \cite{devries2017improved}, AdvBias \cite{cheng2023adversarial}, RandConv \cite{xu2020robust}, and SLAug \cite{su2023rethinking}.
\subsubsection{Quantitative Analysis}
First, we conducted a quantitative analysis. In Table \ref{tab1}, we present the results of the FGML-DG framework on the BraTS dataset, assessing its performance in single segmentation tasks under single-source domain generalization. In Table \ref{tab2}, we display the results of the FGML-DG framework on the abdominal dataset, evaluating its performance in multi-segmentation single-source domain generalization tasks.

\begin{table*}
    \centering
    \caption{Comparison of different methods on the BraTS dataset.}
    \resizebox{\textwidth}{!}{ 
    \begin{tabular}{l|llll|llll|llll|llll}
        \toprule 
         & \multicolumn{8}{c|}{T2}                                                       & \multicolumn{8}{c}{T1CE}                                                      \\ \cline{2-17}
         & \multicolumn{4}{c|}{Dice (\%)}             & \multicolumn{4}{c|}{HD (mm)}          & \multicolumn{4}{c|}{Dice (\%)}             & \multicolumn{4}{c}{HD (mm)}           \\ \midrule 
        Method   & Flair    & T1       & T1CE     & Average  & Flair   & T1      & T1CE    & Average & Flair    & T1       & T2       & Average  & Flair   & T1      & T2      & Average \\ \midrule
        Fed-DG \cite{liu2021feddg}   & 75.77    & 5.82     & 9.51     & 30.37    & 14.45   & 54.03   & 51.06   & 39.85   & 33.03    & 58.30    & 4.09     & 31.72    & 32.07   & 22.35   & 56.08   & 36.83   \\
        MixStyle \cite{zhou2021domain} & 77.03    & 45.68    & 40.23    & 54.31    & 12.97   & 23.10   & 24.36   & 20.14   & 37.55    & 63.12    & 68.31    & 56.32    & 28.75   & 18.74   & 14.91   & 20.80   \\
        CSDG \cite{ouyang2022causality}     & 61.37    & 47.53    & 43.84    & 50.91    & 16.74   & 22.77   & 21.58   & 20.36   & 42.11    & 62.77    & 65.79    & 56.89    & 22.15   & 19.75   & 16.73   & 19.54   \\
        SADN \cite{zhou2022generalizable}     & 75.87    & 49.36    & 38.09    & 54.44    & 13.44   & 20.15   & 23.56   & 19.05   & 47.31    & 63.64    & 63.00    & 57.98    & 21.03   & 18.06   & 17.56   & 18.88   \\
        EGSDG \cite{gutrain}    & 76.16    & 62.43    & 55.87    & 64.82    & 13.43   & 18.84   & 17.79   & 16.69   & 68.49    & 71.22    & \textbf{73.06}    & \textbf{70.92}    & 17.81   & 12.93   & \textbf{14.74}   & 15.16   \\ \midrule
        Ours     & \textbf{77.32}    & \textbf{66.33}    & \textbf{65.25}    & \textbf{69.63}    & \textbf{10.34}   & \textbf{14.42}   & \textbf{15.36}   & \textbf{13.37}   & \textbf{71.79}    & \textbf{71.67}    & 67.93    & 70.46    & \textbf{15.78}   & \textbf{12.14}   & 14.93   & \textbf{14.28}   \\ \bottomrule 
    \end{tabular}
    }
    \label{tab1}
\end{table*}

On the BraTS dataset, our method consistently outperforms or is comparable to the best baseline (EGSDG) in terms of both Dice and HD metrics. Specifically, FGML-DG achieved average Dice scores of 69.63\% and 70.46\% on the two source domains, respectively, with HD scores of 13.37 and 14.28, demonstrating significant superiority over existing state-of-the-art (SOTA) methods.

\begin{table*}
    \centering
    \caption{Comparison of different methods on the Abdominal Multi-Organ dataset.}
    \begin{tabular}{l|lllll|lllll}
        \toprule 
         & \multicolumn{5}{c|}{CT-MRI}                     & \multicolumn{5}{c}{MRI -CT}                     \\ \midrule 
        Method   & Liver  & R-Kidney & L-Kidney & Spleen & Average & Liver  & R-Kidney & L-Kidney & Spleen & Average \\ \midrule
        Cutout \cite{devries2017improved}   & 79.80  & 82.32    & 82.14    & 76.24  & 80.12   & 86.99  & 63.66    & 73.74    & 57.60  & 70.50   \\
        MixStyle \cite{zhou2021domain} & 77.63  & 78.41    & 78.03    & 77.12  & 77.80   & 86.66  & 48.26    & 65.20    & 55.68  & 63.95   \\
        AdvBias \cite{cheng2023adversarial}  & 78.54  & 81.70    & 80.69    & 79.73  & 80.17   & 87.63  & 52.48    & 68.28    & 50.95  & 64.84   \\
        RandConv \cite{xu2020robust} & 73.63  & 79.69    & 85.89    & 83.43  & 80.66   & 84.14  & 76.81    & 77.99    & 67.32  & 76.56   \\
        CSDG \cite{ouyang2022causality}     & 86.62  & 87.48    & 86.88    & 84.27  & 86.31   & 85.62  & 80.02    & 80.42    & 75.56  & 80.40   \\
        SLAug \cite{su2023rethinking}    & \textbf{90.08}  & \textbf{89.23}    & \textbf{87.54}    & \textbf{87.67}  & \textbf{88.63}   & \textbf{89.26}  & \textbf{80.98}    & \textbf{82.05}    & \textbf{79.93}  & \textbf{83.05}   \\ \midrule
        Ours     & 82.24  & 86.43    & 86.86    & 83.21  & 84.69   & 87.21  & 78.24    & 79.32    & 78.29  & 80.77   \\ \bottomrule 
    \end{tabular}
    \label{tab2}
\end{table*}

On the abdominal dataset, our method is only slightly inferior to the current state-of-the-art method SLAug but still demonstrates competitive results. In the CT-to-MRI generalization task, our method achieved an average Dice score of 84.69\%. Although it is slightly inferior to the current state-of-the-art method SLAug (88.63\%), it outperforms or is comparable to other comparative methods, such as RandConv (80.66\%) and CSDG (86.31\%), among others. In the MRI-to-CT generalization task, our method achieved an average Dice score of 80.77\%. This result is superior to CSDG (80.40\%) and significantly surpasses methods such as Cutout, MixStyle, and AdvBias, while only being slightly inferior to SLAug (83.05\%).

Overall, the quantitative experimental results demonstrate that the FGML-DG framework performs excellently in handling multi-modal (BraTS) and cross-domain (abdominal) medical image segmentation tasks. In particular, it achieved leading or highly competitive performance on the BraTS dataset. Additionally, it exhibited competitive results on the more heterogeneous abdominal cross-domain dataset, which validates the robustness of our framework. The slight inferiority of FGML-DG to SLAug on the abdominal dataset may be attributed to the substantial differences in physical imaging principles and feature distributions between CT and MRI. Although our alignment strategy based on style statistics (such as mean and variance) and the memory recall mechanism (MetaStyle) is highly effective in addressing intra-MRI modal differences (as evidenced by the BraTS dataset), it may not fully capture and align all key domain-specific information when confronted with the more complex disparities between CT and MRI. In other words, relying solely on first- and second-order statistics may be insufficient to completely bridge such large domain gaps. In contrast, methods specifically designed for strong domain adaptation tasks (such as SLAug) may hold an advantage in this particular, highly divergent scenario. Nevertheless, the robustness and competitiveness displayed by FGML-DG across datasets fully validate the effectiveness and potential of our proposed Feynman-inspired meta-learning framework in enhancing the model’s domain generalization capabilities.

\subsubsection{Qualitative Analysis}
\begin{figure*}[ht]
  \centering
  \includegraphics[width=0.9\linewidth]{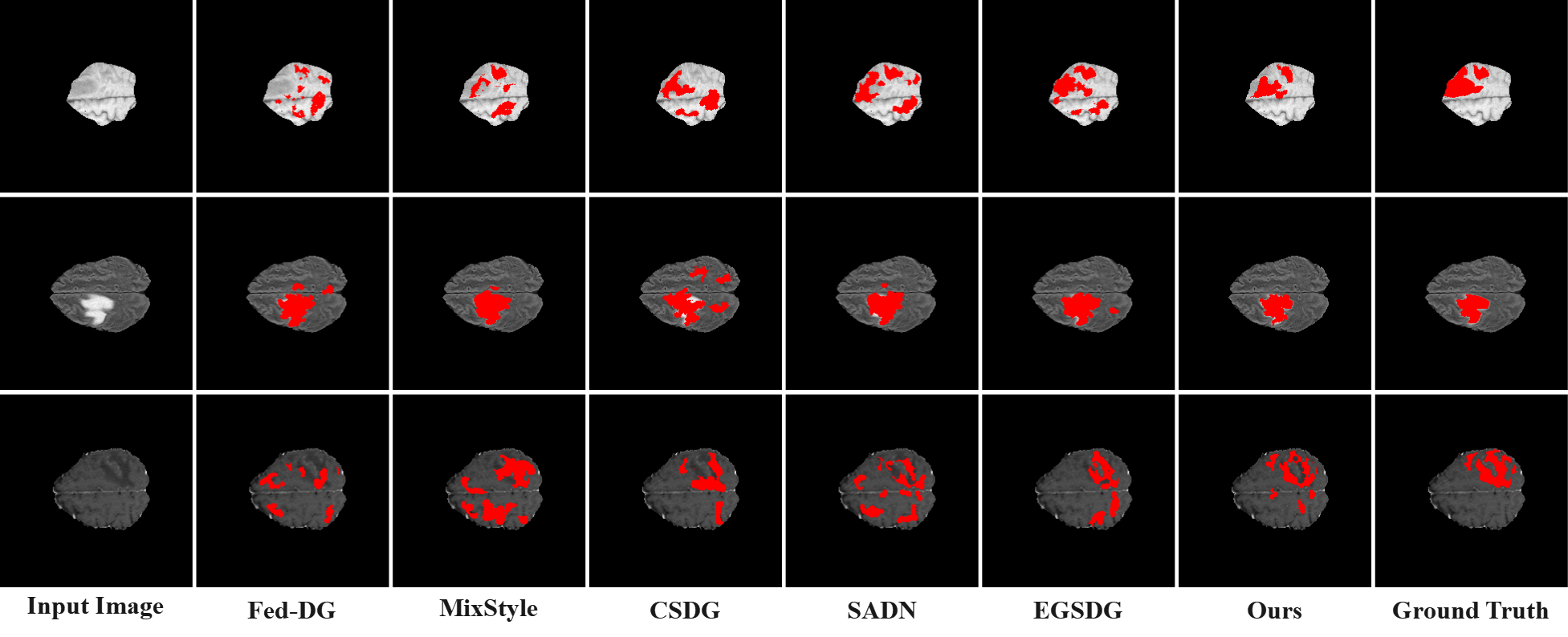}
  \caption{Visual Comparison Results on the BraTS Dataset.}
  \label{fig3}
\end{figure*}

\begin{figure*}[ht]
  \centering
  \includegraphics[width=0.9\linewidth]{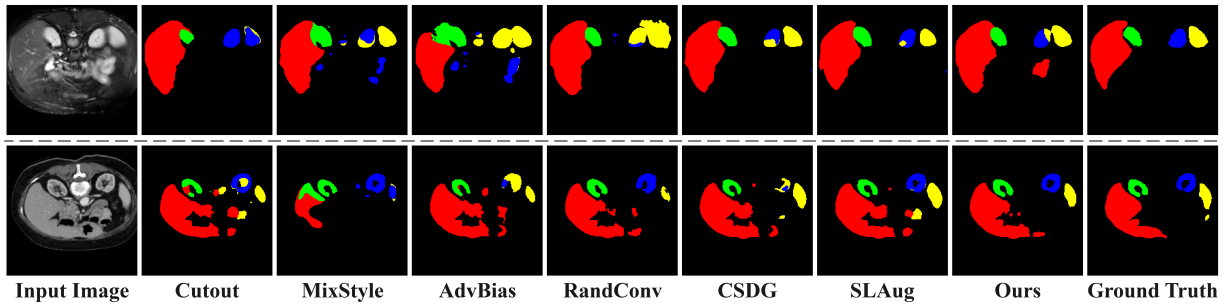}
  \caption{Visual Comparison Results on the Abdominal Multi-Organ Dataset.}
  \label{fig4}
\end{figure*}

We also visualized the segmentation results of our method and other methods on the two tasks in Figures \ref{fig3} and \ref{fig4}. The qualitative results demonstrate that the model achieves accurate segmentation of the target regions.

\subsection{Ablation Experiment}
\subsubsection{Module Ablation Experiment}
To verify the effectiveness and necessity of each key component in our proposed FGML-DG framework, we conducted a series of ablation studies. These experiments aim to independently evaluate the contributions of Meta-Style Knowledge Alignment (MKA), Meta-Style Memory and Review mechanism (MetaStyle), and Feedback-Driven Re-Training strategy (FDRT) to the overall model performance. We selected the BraTS 2018 dataset and used T2 as the source domain for evaluation. As a reference for the lower performance bound, we employed a model that only includes the basic meta-learning framework (denoted as “Meta-Base”). By adding these modules, we quantify the role of each component in improving cross-domain generalization capability. The experimental results are shown in Table \ref{tab3}.

\begin{table}[]
    \centering
    \caption{Ablation Experiment Results for the Key Components of the FGML-DG Framework.}
    \begin{tabular}{l|lll|ll}
        \hline
                  & MKA & MetaStyle & FDRT & Avg.Dice & Avg.HD \\ \hline
        Meta - Base & $\times$ & $\times$ & $\times$ & 25.33    & 45.23  \\ \hline
        Varant 1  & $\times$ & $\checkmark$ & $\times$ & 26.04    & 44.37  \\
        Varant 2  & $\times$ & $\times$ & $\checkmark$ & 28.67    & 41.98  \\
        Varant 3  & $\times$ & $\checkmark$ & $\checkmark$ & 29.97    & 40.03  \\
        Varant 4  & $\checkmark$ & $\times$ & $\times$ & 50.36    & 24.74  \\
        Varant 5  & $\checkmark$ & $\checkmark$ & $\times$ & 60.21    & 16.67  \\
        Varant 6  & $\checkmark$ & $\times$ & $\checkmark$ & 65.56    & 14.29  \\ \hline
        FGML - DG   & $\checkmark$ & $\checkmark$ & $\checkmark$ & \textbf{69.63}    & \textbf{13.37}  \\ \hline
    \end{tabular}
    \label{tab3}
\end{table}

From the results in Table \ref{tab3}, it can be seen that each module we proposed makes a positive contribution to improving the model’s generalization performance. At the same time, we also discovered some interesting phenomena. First, the most significant performance improvement comes from the meta-style knowledge alignment method. When MKA (Variant 4) is added alone to the Meta-Base (with an average Dice score of 25.33\%), the average Dice score jumps substantially to 50.36\%. This substantial improvement highlights the importance of cross-domain feature alignment, particularly the alignment of style information. This aligns closely with the primary step emphasized in the Feynman learning technique—“Concept Understanding.” In our framework, MKA helps the model first “understand” the essential commonalities and differences in style aspects across different domains by simplifying and aligning the style statistics of cross-domain data, thereby laying a solid foundation for subsequent generalization learning. Without effective knowledge alignment, the model struggles to learn transferable knowledge from the source domain. At the same time, the feedback-driven re-training (FDRT) strategy also demonstrates important value. For example, based on the inclusion of MKA (comparing Variant 4 to Variant 6), adding FDRT increases the average Dice score from 50.36\% to 65.56\%. This indicates that targeted re-learning based on feedback from the model on the validation set can effectively help the model overcome weaknesses and consolidate knowledge. This also reflects the idea in the Feynman learning technique of using output feedback to identify weak points and conduct targeted review, which is crucial for enhancing the model’s final generalization capability.

\subsubsection{Ablation Experiment on Knowledge Alignment Loss}
After verifying the overall effectiveness of the meta-style knowledge alignment method, we further conducted an ablation analysis on its internal key loss functions to investigate the specific contributions of the style alignment loss (Eq.\ref{eqalign}) and the prediction consistency loss (Eq.\ref{cons}). The style alignment loss aims to directly minimize the distance between style statistics across different domains through contrastive learning, while the consistency loss enforces the model to produce consistent semantic predictions for inputs from both the source domain and augmented domains. We performed this set of experiments under the same settings as the aforementioned ablation studies (i.e., the BraTS 2018 dataset with the T2 source domain). Using the complete FGML-DG model as the baseline, we evaluated the impact on final performance by removing $L_{align}$ or $L_{cons}$ individually, as well as removing both simultaneously.

\begin{table}
    \centering
    \caption{Ablation Experiment Results for the Loss Functions in the Meta-Style Knowledge Alignment Method.}
    \resizebox{\linewidth}{!}{ 
    \begin{tabular}{l|ll|ll}
        \hline
                                 & $L_{align}$ & $L_{cons}$ & Avg.Dice & Avg.HD \\ \hline
        FGML-DG w/o $L_{cons}$ + $L_{align}$ & $\times$ & $\times$ & 29.97    & 40.03  \\
        FGML-DG w/o $L_{align}$          & $\times$ & $\checkmark$ & 40.21    & 32.57  \\
        FGML-DG w/o $L_{cons}$           & $\checkmark$ & $\times$ & 60.33    & 19.38  \\
        FGML-DG                      & $\checkmark$ & $\checkmark$ & \textbf{69.63}    & \textbf{13.37}  \\ \hline
    \end{tabular}
    }
    \label{tab4}
\end{table}

The results in Table \ref{tab4} clearly demonstrate the individual contributions of the style alignment loss ($L_{align}$) and the consistency loss ($L_{cons}$) within the meta-style knowledge alignment component. Compared to the complete FGML-DG model (Avg. Dice 69.63\%, Avg. HD 13.37): removing the consistency loss ($L_{cons}$) alone results in a decline in average Dice to 60.33\% and an increase in average HD to 19.38. This indicates that enforcing semantic consistency in predictions between the source domain and augmented domains is beneficial for improving segmentation accuracy and boundary precision. Removing the style alignment loss ($L_{align}$) alone leads to a more substantial performance degradation, with average Dice dropping to 40.21\% and average HD increasing significantly to 32.57. This significant difference highlights that directly aligning style feature statistics across different domains through contrastive learning ($L_{align}$) is the most critical component in the meta-style knowledge alignment method.

\section{Conclusion}
In this work, we introduce the FGML-DG framework, a novel meta-learning method inspired by Feynman learning techniques derived from cognitive science, aimed at addressing the key challenges of domain generalization in medical image segmentation across multiple modalities. By integrating principles such as conceptual understanding, knowledge reuse, and feedback-driven optimization, our method significantly enhances the model’s adaptability and generalization performance. It outperforms state-of-the-art methods on the BraTS 2018 multi-modal dataset, achieving higher Dice scores and lower Hausdorff distances. Competitive results are also obtained on the more heterogeneous abdominal cross-domain dataset. These findings highlight the effectiveness of mimicking human cognitive processes in AI. The FGML-DG framework not only advances domain generalization in medical imaging but also bridges cognitive science and AI, offering a promising paradigm for developing more robust and interpretable models for medical applications.





\bibliography{mybibfile}

\end{document}